\documentclass[letterpaper]{article} 
\usepackage{aaai23}  
\usepackage{times}  
\usepackage{helvet}  
\usepackage{courier}  
\usepackage[hyphens]{url}  
\usepackage{graphicx} 
\usepackage{natbib}  
\usepackage{caption} 
\usepackage{multirow}
\usepackage{booktabs}
\usepackage{subfigure}
\usepackage{amsmath,amssymb}
\usepackage{algorithm}
\usepackage{algorithmic}
\begin{document}

%
\title{Data Imputation with Iterative Graph Reconstruction}
\author{
Jiajun Zhong, Ning Gui\thanks{Corresponding author.}, Weiwei Ye\\
}

\affiliations{
    School of Computer Science and Engineering, Central South University, China \\
    214711110@csu.edu.cn, ninggui@csu.edu.cn, 634948676@qq.com \\
}

\maketitle
\begin{abstract}

  
  Effective data imputation demands rich latent ``structure" discovery capabilities from ``plain" tabular data. Recent advances in graph neural networks-based data imputation solutions show their structure learning potentials by translating tabular data as bipartite graphs. However, due to a lack of relations between samples, they treat all samples equally which is against one important observation: ``similar sample should give more information about missing values."  This paper presents a novel Iterative graph Generation and Reconstruction framework for Missing data imputation(IGRM). Instead of treating all samples equally, we introduce the concept: ``friend networks" to represent different relations among samples. To generate an accurate friend network with missing data, an end-to-end friend network reconstruction solution is designed to allow for continuous friend network optimization during imputation learning. The representation of the optimized friend network, in turn, is used to further optimize the data imputation process with differentiated message passing. Experiment results on eight benchmark datasets show that IGRM yields 39.13\% lower mean absolute error compared with nine baselines and 9.04\% lower than the second-best. Our code is available at \url{https://github.com/G-AILab/IGRM}.

\end{abstract}

\section{Introduction}


Due to diverse reasons, missing data becomes a ubiquitous problem and  It can introduce an amount of bias and make most machine learning methods inapplicable~\cite{bertsimas2017predictive}. Missing data has been tackled by direct deletion or by data imputation. Deletion-based methods discard incomplete samples and result in the loss of important information especially when the missing ratio is high~\cite{luo2018multivariate}. Data imputation is more plausible as it estimates missing data with values based on some statistical or machine learning techniques~\cite{lin2020missing} and allows downstream machine learning methods normal operation with the ``completed" data~\cite{muzellec2020missing}.

Existing imputation methods discover latent relations in tabular data from different angles and normally with diverse assumptions.
Simple statistical methods are feature-wise imputation, ignoring relations among features and might bias downstream tasks\cite{jager2021benchmark}.
Machine learning methods such as k-nearest neighbors~\cite{troyanskaya2001missing} and matrix completion~\cite{hastie2015matrix} can capture higher-order interactions between data from both sample and feature perspectives. However, they normally make strong assumptions about data distribution or missing data mechanisms and hardly generalize to unseen samples~\cite{you2020handling}.
Some solutions, e.g. MIRACLE~\cite{kyono2021miracle}, try to discover causal relations between features which are only small parts of all possible relations. GINN~\cite{spinelli2020missing} discovers relations between samples while failing to model relations between samples and features. These methods typically fall short since they do not employ a data structure that can simultaneously capture the intricate relations in features, samples, and their combination.



A graph is a data structure that can describe relationships between entities. It can model complex relations between features and samples without restricting predefined heuristics. Several Graph Neural Networks(GNNs)-based approaches have been recently proposed for graph-based data imputation. Those approaches generally transform tabular data into a bipartite graph with samples and features viewed as two types of nodes and the observed feature values as edges~\cite{berg2017graph,you2020handling}. 
However, due to a lack of sample relations, those solutions commonly treat all samples equally and only rely on different GNN algorithms to learn this pure bipartite graph. They fail to use one basic observation, ``similar sample should give more information about missing values". The observation, we argued, demands us to differentiate the importance of samples during the graph learning process. 

\begin{figure}[btp]
\centering
\includegraphics[width=1.0\columnwidth]{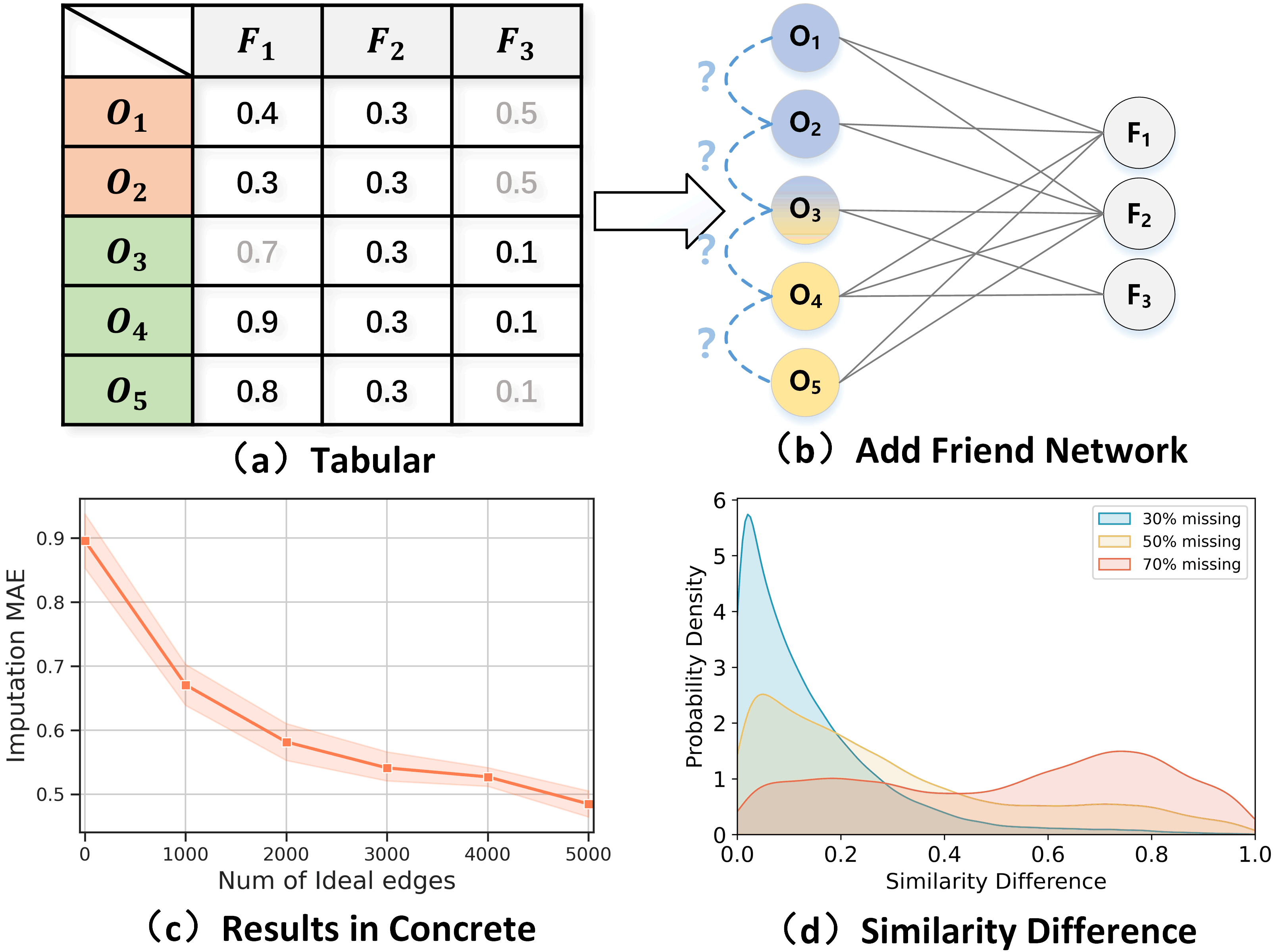}
\caption{Imputation with friend network. Gray indicates missing values. Similar samples are marked with the same color with the ground-truth in a) and missing data in b). c) shows the trend of IGRM with adding an ideal friend network on Concrete in UCI dataset. d) shows similarity difference between ground-truth and data with missing.}
\label{fig:intro}
\end{figure}

Fig.~\ref{fig:intro} highlights the importance and difficulty in using the latent samples relations for data imputation. The tabular data with some missing data(Fig.~\ref{fig:intro}a) can be easily transformed into a bipartite graph (Fig.~\ref{fig:intro}b) and learned with GNNs solutions. However, if we know the ground-truth samples, we can form so-called \textit{friend networks} by connecting similar samples with additional edges, shown in Fig.~\ref{fig:intro}b. If we can fully exploit the friend network(by IGRM), the MAE of data imputation significantly drops, as shown in Fig.~\ref{fig:intro}c. It clearly shows the importance of adopting the friend network. However, the large portion of missing data makes it hard to acquire accurate relations among samples. The cosine similarity on samples with missing data can be significantly biased, as shown in Fig.~\ref{fig:intro}d, especially when the missing ratio is above 50\%. Thus, a new mechanism is needed to learn accurate sample relations under a high missing ratio.


This paper proposes IGRM, an end-to-end Iterative graph Generation and Reconstruction framework for Missing data imputation. IGRM goes beyond the bipartite graph with the equal sample used in \cite{you2020handling} by using the \textit{friend network} to differentiate the bipartite graph learning process. IGRM also addresses the problem of how to generate an accurate friend network under a high missing ratio with an end-to-end optimization solution. 

\noindent\textbf{Contributions.} 1) We propose a novel usage of ``friend network" in data imputation. The introduction of the friend network augments traditional bipartite graphs with capabilities to differentiate sample relations. 2) To build an accurate friend network under missing data, IGRM is designed to be end-to-end trainable. The friend network is continuously optimized by the imputation task and is used to further optimize the bipartite graph learning process with differentiated message passing. 3) Instead of using plain attribute similarity, we innovatively propose the use of node embedding to reduce the impacts of a large portion of missing data and diverse distributions of different attributes. 4) We compared IGRM with nine state-of-the-art baselines on eight benchmark datasets. Compared with the second-best baselines, IGRM yields 9.04\% lower mean absolute error (MAE) at the 30\% data missing rate and even higher improvements in data with higher missing rates. A set of ablation studies also prove the effectiveness of our proposed designs. 

\section{Related Work}
Studies on data imputation can be generally classified into the following three streams. 


\subsubsection{Statistics-based} include single imputations and multiple imputations.
Single imputations are calculated only with the statistical information of non-missing values in that feature dimension~\cite{little1987statistical,acuna2004treatment,donders2006gentle}, leading to possible distorted results due to biased sample distributions.
Multivariate imputation by chained equations(MICE)~\cite{buuren2010mice,white2011multiple,azur2011multiple} performs data imputation by combing multiple imputations. However, it still relies on the assumption of missing at random(MAR)~\cite{little1987statistical}. Beyond the MAR assumption, MICE might result in biased predictions~\cite{azur2011multiple}.

\subsubsection{Machine and Deep-learning based} have been developed to address imputation, including k-nearest neighbors(KNN), matrix completion, deep generative techniques, optimal transport, and others. 
KNNs~\cite{troyanskaya2001missing,keerin2012cluster,malarvizhi2012k} are limited in making weighted averages of similar feature vectors, while matrix completions\cite{hastie2015matrix,genes2016recovering,fan2020polynomial} are transductive and cannot generalize to unseen data. 
Deep generative techniques include denoising auto-encoders(DAE)~\cite{vincent2008extracting,nazabal2020handling,gondara2017multiple} and generative adversarial nets(GAN)~\cite{yoon2018gain,allen2016generative}. They demand complete training datasets and may introduce bias by initializing missing values with default values. 
OT~\cite{muzellec2020missing} assumes that two batches extracted randomly in the same dataset have the same distribution and uses the optimal transport distance as a training loss. Miracle~\cite{kyono2021miracle} encourages data imputation to be consistent with the causal structure of the data by iteratively correcting the shift in distribution. However, those approaches fail to exploit the potential complex structures in and/or between features and samples.

\subsubsection{Graph-based. } Several graph-based solutions are proposed with state-of-art data imputation performances. 
Both GC-MC~\cite{berg2017graph} and IGMC~\cite{zhang2019inductive} use the user-item bipartite graph to achieve matrix completion. However, they  assign separate processing channels for each rating type and fail to generalize to the continuous data. To address this limitation, GRAPE~\cite{you2020handling} adopts an edge learnable GNN solution and can handle both continuous and discrete values. However, its pure bipartite graph solution largely ignores potentially important contributions from similar samples. 
GINN~\cite{spinelli2020missing} exploits the euclidean distance of observed data to generate a graph and then reconstruct the tabular by graph denoising auto-encoder.

\section{Preliminaries} 
Let data matrix $\mathbf{D}\in\mathbf{R}^{n \times m}$ indicate the tabular with missing data, consisting of $n$ samples and $m$ features. 
Let $\mathbf{U}=\{u_1,...,u_n\}$ and $\mathbf{V} = \{v_1,...v_m\}$ be the sets of samples and features respectively.
The $j$-th feature of the $i$-th samples is denoted as $\mathbf{D}_{ij}$. 
A binary mask matrix $\mathbf{N}\in\mathbf\{0,1\}^{n\times m}$ is given to indicate the location of missing values in tabular, where $\mathbf{D}_{ij}$ is missing only if $\mathbf{N}_{ij} = 0$. In this paper, the goal of data imputation is to predict the missing data 
point $\hat{\mathbf{D}}_{ij}$ at  $\mathbf{N}_{ij}=0$.


\noindent\textbf{Definition 1. } \textit{Data matrix as a bipartite graph.}
Data matrix $\mathbf{D}$ can be formulated as a undirected bipartite graph $\mathbf{\mathcal{B}}=(\mathbf{U},\mathbf{V},\mathbf{E})$.
$\mathbf{E}$ is the edge set where $\mathbf{E}_{ij}$ is an edge between $u_i$ and $v_j$: $\mathbf{E} = \{(u_i,v_j,{e}_{ij})|u_i \in \mathbf{U}, v_j \in \mathbf{V}, \mathbf{N}_{ij} = 1\}$ where the weight of the edge equals the value of the $j^{th}$ feature of sample i, $e_{ij}$ = $\mathbf{D}_{ij}$. Here, nodes in $\mathbf{U},\mathbf{V}$ do not naturally come with features. Similar to the settings of GRAPE, let vector consist of vector \textbf{1} in $m$ dimension as $\mathbf{U}$ node features and $m$-dimensional one-hot vector as $\mathbf{V}$ node features. 



\noindent\textbf{Definition 2. } \textit{Friend network.}
$\mathcal{F} = (\mathbf{U},\mathbf{A})$ represents similarity among the samples $\mathbf{U}$ by connecting similar samples denoted with a binary adjacency matrix $\mathbf{A}\in \{0,1\}^{n\times n}$. For any two samples $u_i, u_j \in \mathbf{U}$, if they are similar, $\mathbf{A}_{ij}=1$. Here, let $\mathcal{N}(i)$ denotes function that returns the set of samples whom $u_i$ directly connects with in $\mathcal{F}$. 





\noindent\textbf{Definition 3. } \textit{Cosine similarity with missing data.}
With the existence of missing values, the cosine similarity between two samples is defined as follows:
\begin{equation}
    \begin{aligned}
        S_{ij}=cos(\mathbf{D}_{i:}\odot(\mathbf{N}_{i}\odot \mathbf{N}_{j}), \mathbf{D}_{j:}\odot(\mathbf{N}_{i}\odot \mathbf{N}_{j}))
    \end{aligned}
\label{equ:attri_similarity}
\end{equation}
where $\odot$ stands for the Hadamard product between vectors, $cos$ is the cosine similarity function. It computes pairwise similarity for all features, but only non-missing elements of both vectors are used for computations. 
As illustrated in Fig.~\ref{fig:intro}d, this formulation might introduce a large bias since it is calculated on the subspace of non-missing features. When $\mathbf{N}_{i}\odot \mathbf{N}_{j}=\varnothing$, the calculated similarity would be \textit{NaN} and is hard to be used for similarity evaluation.

\noindent\textbf{Definition 4. } \textit{Sample embedding-based similarity.} 
The similarity of two samples $u_i$ and $u_j$ can be calculated with their embeddings $\mathbf{p}_i$ and $\mathbf{p}_j$:
\begin{equation}
S_{ij}^{'}=cos(\mathbf{p}_{i}, \mathbf{p}_{j})    
\end{equation}

Here, the sample embedding $\mathbf{p}_i$ can represent the observed values of the $i$-th sample. As it is generated from structural and feature dimensions, the embedding-based similarity calculation is more reliable than Eq.~\ref{equ:attri_similarity}.


\begin{figure*}[tpb]
    \centering
    \includegraphics[width=0.8\textwidth]{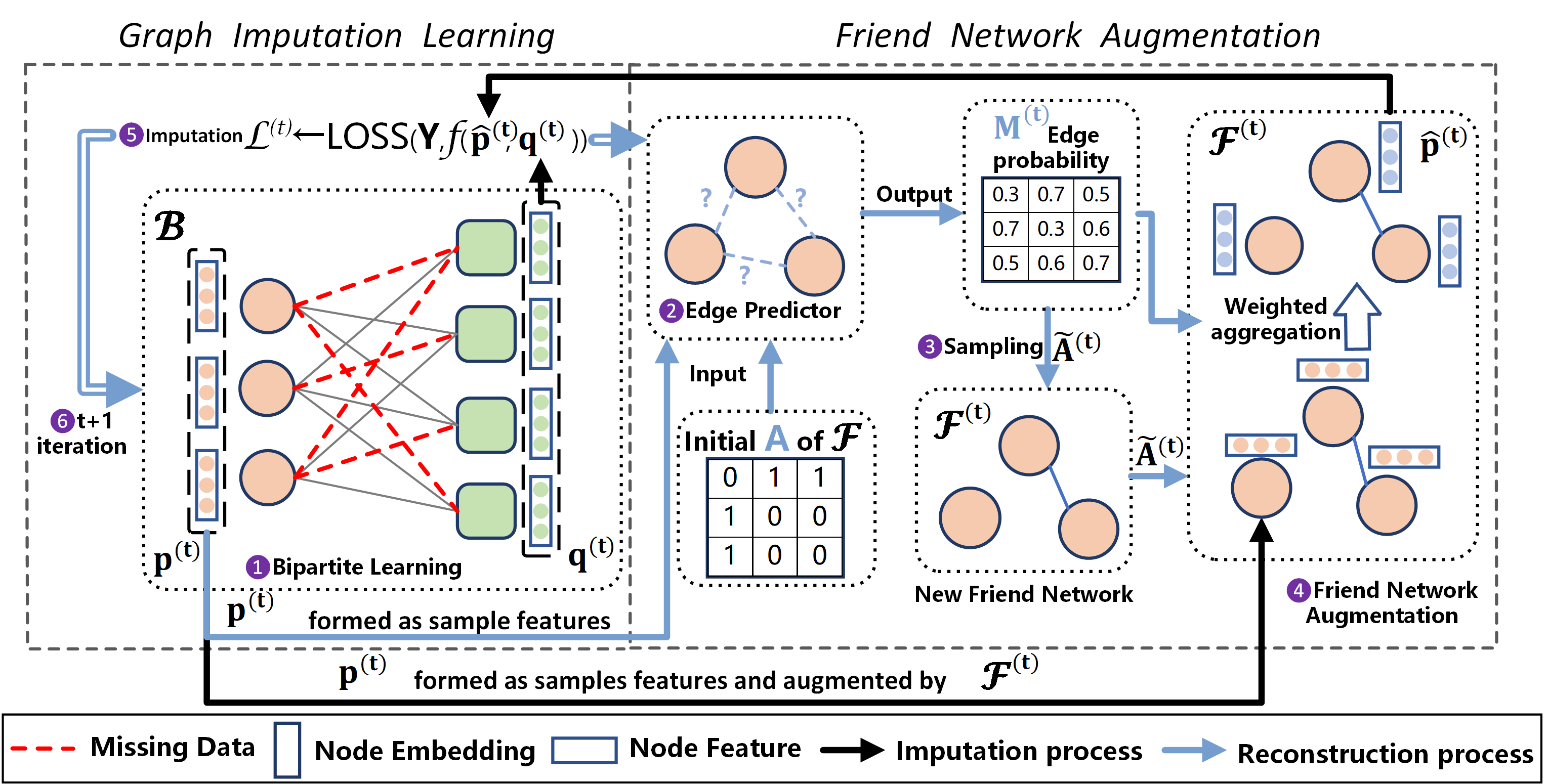}
    \caption{Overall architecture of the proposed IGRM framework in $t$-th iteration.}
    \label{fig:model}
\end{figure*}

\section{IGRM}


This section shows the basic structure of IGRM and illustrates designs that allow for the end-to-end learning of an accurate friend network with missing data. 



\subsection{Network Architecture}

Fig.\ref{fig:model} shows the overall architecture and training process at $t$-th iteration. IGRM is comprised of two modules: the graph imputation learning module for bipartite graph $\mathcal{B}$ learning (left part of Fig.\ref{fig:model}) and the friend network augmentation module (right part). Existing GNNs solutions only have a graph imputation learning module and learn the data imputation model with an edge-level prediction function ${f}$ in the bipartite graph $\mathcal{B}$. The $\hat{D}_{ij} = f_{ij}{(\mathcal{B})}$ by minimizing the difference between $\hat{D}_{ij}$ and ${D}_{ij}$.
In comparison, IGRM introduces a novel friend network augmentation module that does not exist in other Graph-based imputation solutions. 

In IGRM, the key problem lies in how to generate an accurate friend network under the distribution deviation introduced by the missing data and how can the bipartite graph $\mathcal{B}$ and the friend network $\mathcal{F}$ effectively integrate knowledge from each other. Here, we use Graph Representation Learning (GRL)~\cite{hamilton2017representation} to distill important graph knowledge into dense embeddings from the friend network $\mathcal{F}$. Here, let $\hat{\mathbf{p}}$ represent the sample embeddings distilled from the friend network. Then, the bipartite graph learning can be changed from only the bipartite graph $\mathcal{B}$ to the combination of both $\mathcal{B}$ and $\hat{\mathbf{p}}$.

\begin{equation}
    \begin{aligned}
       \hat{f}_{ij}(\mathbf{\mathcal{B}}, \hat{\mathbf{p}})\rightarrow \hat{\mathbf{D}}_{ij} \ \ \forall N_{ij}=0
    \end{aligned}
    \label{eq:IGRMF}
\end{equation}

On the other hand, $\mathcal{F}$ can be optimized with the support of $\mathcal{B}$. Here, let  ${\mathbf{p}}$ represent the sample embeddings distilled from the bipartite graph $\mathcal{B}$. As $\mathbf{p}$ presents the high-level abstraction of samples and can be used as an accurate base for the friend network (re)construction. We can get better ${\mathbf{p}}$ during the GRL training process. We define a function that takes the updated $\mathbf{p}^{(t)}$ to generate an optimized new friend network $\mathbf{\widetilde{A}}^{(t)}$. The friend network augmentation includes network structure augmentation and node feature augmentation. Of course, to avoid possible bias introduced by the missing data, we randomly initialize $\mathcal{F}$ with arbitrary numbers of new edges between samples. The initialization process can generally be replaced with any suitable method. Here, IGRM can be degraded to an existing solution with no relation discovery, e.g., GRAPE, or with one-round relation discovery, e.g., GINN.

\subsection{Bipartite Graph Learning}

In the bipartite graph, the edges of the bipartite graph carry important non-missing value information. Here, we follow the G2SAT~\cite{you2019g2sat} to learn the bipartite graph with a variant of GraphSAGE~\cite{hamilton_grapshsage}, which can utilize the edge attributes. The source node embedding $\mathbf{p_i}$($\mathbf{q_j}$) and edge embedding $\mathbf{e}_{ij}$ are concatenated during message passing:
\begin{equation}
    \begin{aligned}
       \mathbf{h}^{(l)}_{i} \leftarrow Mean(\sigma(\mathbf{W}^{(l)} \cdot Concat(\mathbf{q}^{(l-1)}_{j},\mathbf{e}^{(l-1)}_{ij})
    \end{aligned}
    \label{eq:bipartite_neighbor}
\end{equation}
where $l$ indicates the $l$-th layer, $Mean$ denotes the mean operation with a non-linear transformation $\sigma$, $Concat$ is an operation for concatenation, and $\mathbf{W}$ is the trainable weight. Node embedding is updated by:
\begin{equation}
    \begin{aligned}
        \mathbf{p}^{(l)}_{i}\leftarrow \sigma(\mathbf{\widehat{W}}^{(l)} \cdot Concat(\mathbf{p}^{(l-1)}_{i},\mathbf{h}^{(l)}_{i}))
    \end{aligned}
    \label{eq:bipartite_node}
\end{equation}
where $\mathbf{\widehat{W}}$ is trainable. Edge embedding is updated by:
\begin{equation}
    \begin{aligned}
       \mathbf{e}^{(l)}_{ij} \leftarrow \sigma(\mathbf{Q}^{(l)} \cdot Concat(\mathbf{e}^{(l-1)}_{ij}, \mathbf{p}^{(l)}_{i}, \mathbf{q}^{(l)}_{j})
    \end{aligned}
    \label{eq:edge}
\end{equation}
where $\mathbf{Q}$ is the trainable weight. Eq.\ref{eq:bipartite_neighbor}, \ref{eq:bipartite_node} and \ref{eq:edge} are shown as step-1 in Fig.\ref{fig:model}. Let $\mathbf{p}^{(t)}$ and $\mathbf{q}^{(t)}$ be the sample and feature embeddings at the last layer of $t$-th iteration.

\subsection{Friend Network Augmentation \& Learning}

During training, the friend network should be continuously optimized from both the structure and the feature perspectives. Then, the optimized friend network should also be encoded to support the bipartite graph learning.

\subsubsection{Augmentation.} 

From the bipartite GRL module, we can obtain the sample embeddings $\mathbf{p}^{(t)}$ for $\mathcal{F}$. These embeddings carry rich information and to some extent, can mitigate the impacts of missing data. Thus, this information is being used to guide the friend network (re)construction process for both feature and structure augmentation. 

\noindent\textbf{Feature Augmentation.}
Instead of using the \textbf{1} as features in GRAPE, IGRM uses the vector $\mathbf{p}^{(t)}$ as the feature of friend network $\mathcal{F}$. Let $\hat{\mathbf{X}}^{(t)}$ as the feature matrix of $\mathcal{F}$, we have
\begin{equation}
    \begin{aligned}
       \hat{\mathbf{X}}^{(t)} = \mathbf{p}^{(t)}
    \end{aligned}
    \label{eq:embd2fea}
\end{equation}

\noindent\textbf{Differentiable Structure Augmentation.}
Here, we propose a novel structure reconstruction component that can continuously optimize the friend network structure according to $\mathbf{p}^{(t)}$. Here, we use Graph auto-encoder (GAE)~\cite{kipf2016variational} to learn class-homophilic tendencies in edges~\cite{zhao2021data} with an edge prediction pretext task. GAE takes initial $\mathbf{A}$ and $\hat{\mathbf{X}}^{(t)}$ as input and outputs an edge probability matrix $\mathbf{M}^{(t)}$. Of course, GAE can generally be replaced with any suitable models. 
\begin{equation}
    \begin{aligned}
       \mathbf{M}^{(t)}=\sigma(\mathbf{Z}\mathbf{Z}^{T}), \mathbf{Z}=GCN_{1}(\mathbf{A},GCN_{0}(\mathbf{A},\hat{\mathbf{X}}^{(t)}))
    \end{aligned}
    \label{eq:GAE}
\end{equation}
where $\mathbf{Z}$ is the hidden embedding learned from the encoder, $\sigma$ is the activation function. Then, $\mathbf{M}^{(t)}$ is the edge probability matrix established by the inner-product decoder. The higher the $M_{ij}$, the higher possibility the sample i and j are similar. Thus, we can use $\mathbf{M}^{(t)}$ to generate the new structure of $\mathcal{F}^{(t)}$. 
Eq.\ref{eq:GAE} is shown as step-2 in Fig.~\ref{fig:model}.


However, the sampling process of edges from $\mathbf{M}^{(t)}$, such as $argmax(\mathbf{M}^{(t)})$, would cause interruption of gradient back propagation since samples from a distribution on discrete objects are not differentiable with respect to the distribution parameters\cite{kusner2016gans}.
Therefore, for trainable augmentation purpose, we employ Gumbel-softmax reparameterization\cite{jang2016categorical,maddison2016concrete} to obtain a differentiable approximation by building continuous relaxations of the discrete parameters involved in the sampling process.
\begin{equation}
    \begin{aligned}
       \mathbf{\widetilde{A}}^{(t)} = \frac{exp((\log(\mathbf{M}^{(t)}_{ij})+g_i)/\tau}{\sum^{k}_{j=i}exp((\log(\mathbf{M}^{(t)}_{ij})+g_j)/\tau)}
    \end{aligned}
    \label{eq:gumbel}
\end{equation}
where $g_i\cdots g_k$ are independent and identically distributed samples drawn from Gumbel(0,1) and $\tau$ is the softmax temperature. 
The friend network is reconstructed as $\mathbf{\mathcal{F}}^{(t)}=(\mathbf{U}, \mathbf{\widetilde{A}}^{(t)})$.
Eq.\ref{eq:gumbel} is shown as step-3.


\subsubsection{Representation Learning}
In order to let similar samples contribute more useful information, we need to encode the friend network into embeddings so $\mathcal{F}$ can be easily used for the bipartite graph imputation learning. Here, we adopt the well-known GraphSAGE on $\mathbf{\mathcal{F}}^{(t)}$ for representation learning in which nodes use $\mathbf{M}^{(t)}$ to perform weighted aggregation of neighbor information.
It takes $\hat{\mathbf{X}}^{(t)}$, $\mathbf{\widetilde{A}}^{(t)}$ and $\mathbf{M}^{(t)}$ as input and outputs the embeddings $\hat{\mathbf{p}}^{(t)}$:
\begin{equation}
    \begin{split}
      & \hat{\mathbf{p}}^{(t)}_\mathbf{i}\leftarrow \delta(\mathbf{O} \cdot Concat(\mathbf{p}^{(t)}_{i},\sum_{u_j\in \mathcal{N}(i)}\hat{\mathbf{X}}^{(t)}_{j}\mathbf{M}^{(t)}_{ij}))
    \end{split}
    \label{eq:friend_augment}
\end{equation}
where $\mathbf{O}$ is the trainable weight, $\delta$ is the activation function. Eq.\ref{eq:friend_augment} is shown as step-4.

\subsection{Imputation Task} 
Missing values are imputed with corresponding node embeddings $\hat{\mathbf{p}}^{(t)}$ and $\mathbf{q}^{(t)}$:
\begin{equation}
    \begin{aligned}
       \mathbf{Y}_{pred}=\sigma(f(Concat(\hat{\mathbf{p}}^{(t)},\mathbf{q}^{(t)})))
    \end{aligned}
    \label{eq:predict}
\end{equation}
We then optimize models with the edge prediction task. The imputation loss $\mathcal{L}$ is:
\begin{equation}
    \begin{aligned}
       \mathcal{L}=CE(\mathbf{Y}_{pred},\mathbf{Y}) + MSE(\mathbf{Y}_{pred},\mathbf{Y})
    \end{aligned}
    \label{eq:loss}
\end{equation}
where $\mathbf{Y}=\{\mathbf{D}_{ij}| \forall \mathbf{N}_{ij}=0\}$, we use cross-entropy (CE) when imputing discrete values and use mean-squared error (MSE) when imputing continuous values. Eq.\ref{eq:predict} and \ref{eq:loss} are shown as step-5.
The full process is illustrated in Algorithm \ref{alg:IGRM}.
\begin{algorithm}[htbp]
\caption{General Framework for IGRM}
\label{alg:IGRM}
\textbf{Input}: $\mathbf{\mathcal{B}}=(\mathbf{U},\mathbf{V},\mathbf{E})$, $\mathbf{N}$, $\mathbf{A}$, iteration T\\
\textbf{Output}: $\hat{\mathbf{Y}}$, $\widetilde{\mathbf{A}}$
\begin{algorithmic}[1] 
\STATE $\mathbf{X}\leftarrow$ \textbf{1} \textbf{if} continuous \textbf{else} ONEHOT
\STATE $t\gets 0$
\WHILE{$t < T$}
    \STATE $\mathbf{p}^{(t)},\mathbf{q}^{(t)}\leftarrow GNN_{1}(\mathbf{X},\mathbf{E})$ using Eq.\ref{eq:bipartite_neighbor}, \ref{eq:bipartite_node} and \ref{eq:edge}
    \STATE $\hat{\mathbf{X}}^{(t)} = \mathbf{p}^{(t)}$
    \IF {reconstruct}
        \STATE $\mathbf{M}^{(t)}\leftarrow GAE(\mathbf{A},\hat{\mathbf{X}}^{(t)})$ using Eq.\ref{eq:GAE}
        \STATE $\widetilde{\mathbf{A}}^{(t)}\leftarrow Sampling(\mathbf{M}^{(t)})$ using Eq.\ref{eq:gumbel}
    \ENDIF
    \STATE $\hat{\mathbf{p}}^{(t)}\leftarrow GNN_{2}(\hat{\mathbf{X}}^{(t)},\widetilde{\mathbf{A}}^{(t)}, \mathbf{M}^{(t)})$ using Eq.\ref{eq:friend_augment} 
    \STATE $\mathbf{Y}_{pred}=\sigma(f(CONCAT(\hat{\mathbf{p}}^{(t)},\mathbf{q}^{(t)})))$
    \STATE $\mathcal{L}=CE(\mathbf{Y}_{pred},\mathbf{Y}) + MSE(\mathbf{Y}_{pred},\mathbf{Y})$
    \STATE $t\gets t+1$
\ENDWHILE
\end{algorithmic}
\end{algorithm}

\section{Experiments}
\subsection{Experimental Setup}
\subsubsection{Dataset}
We evaluate IGRM on eight real-world datasets from the UCI Machine Learning repository~\cite{asuncion2007uci} and Kaggle\footnote{https://www.kaggle.com/datasets}. The first four datasets come from UCI and others come from Kaggle.
These datasets consist of mixed data types with both continuous and discrete variables and cover different domains:  biomedical(Heart, Diabetes), E-commerce, finance(DOW30), etc. The summary statistics of these datasets are shown in Table \ref{tab:datasets}. 

\begin{table}[hbtp]
 \centering
  \caption{Properties of eight datasets, Disc. short for discrete, Cont. short for continuous, and F. short for features.}
    \begin{tabular}{lrrr}
    \toprule
    \textbf{Dataset}  & \textbf{Samples \#} & \textbf{Disc. F. \#} & \textbf{Cont. F. \#} \\
    \midrule
    Concrete & 1030  & 0 & 8 \\
    Housing & 506  & 1 & 12 \\
    Wine  &  1599  & 0 & 11 \\
    Yacht & 308  & 0 & 6 \\
    Heart & 1025  & 9 & 4 \\
    DOW30 & 2448  & 0 & 12 \\
    E-commerce & 10999  & 7 & 3 \\
    Diabetes & 520  & 15 & 1 \\
    \bottomrule
    \end{tabular}
  \label{tab:datasets}
\end{table}%

\subsubsection{Baseline models}
We compare IGRM against nine commonly used imputation methods, including statistical methods and machine learning models. The implementations used are mainly from respective papers. 
Mean: a feature-wise mean imputation. 
KNN~\cite{hastie2015matrix}: the mean of the K-nearest observed neighbors as imputation. 
MICE~\cite{buuren2010mice}: provides multiple imputation values for each missing value with different missing assumptions. 
SVD~\cite{troyanskaya2001missing}: imputes data by performing single value decomposition on the dataset.
Spectral~\cite{mazumder2010spectral}: imputes data  with a soft-thresholded SVD.
GAIN~\cite{yoon2018gain}: imputes data using Generative Adversarial Nets. 
OT~\cite{muzellec2020missing}: a deep neural distribution matching method based on optimal transport (OT). Here, the Sinkhorn variant is chosen.
Miracle~\cite{kyono2021miracle}:  a causally-aware imputation framework.
GINN~\cite{spinelli2020missing}:  an imputation framework based on the graph denoising autoencoder.
GRAPE~\cite{you2020handling}:a pure bipartite graph-based framework for feature imputation.
    
\begin{table*}[thp]
 \centering
  \caption{Comparisons of averaged(*10) MAE values for missing data imputation on eight datasets with 30\% and 70\% MCAR. Results are enlarged by 10 times. Results for MAR and MNAR hold similar trends and can be found in Appendix}
    \resizebox{1.0\textwidth}{!}{
    \begin{tabular}{l|cc|cc|cc|cc|cc|cc|cc|cc}
    \toprule
    \textbf{Dataset} & \multicolumn{2}{c|}{\textbf{Concrete}} & \multicolumn{2}{c|}{\textbf{Housing}} & \multicolumn{2}{c|}{\textbf{Wine}} & \multicolumn{2}{c|}{\textbf{Yacht}} & \multicolumn{2}{c|}{\textbf{Heart}} & \multicolumn{2}{c|}{\textbf{DOW30}} & \multicolumn{2}{c|}{\textbf{E-commerce}} & \multicolumn{2}{c|}{\textbf{Diabetes}} \\
    \textbf{Ratio} & 30\% & 70\% & 30\% & 70\% & 30\% & 70\% & 30\% & 70\% & 30\% & 70\% & 30\% & 70\% & 30\% & 70\% & 30\% & 70\%\\
    \midrule
    Mean & 1.83 & 1.81 & 1.82 & 1.82 & 0.98 & 0.98 & 2.14 & \underline{2.18} & 2.33 & 2.33 & 1.53 & 1.53 & 2.51 & \underline{2.51} & 4.37 & 4.35 \\
    KNN & 1.25 & 1.74 & 0.96 & 1.52 & 0.80 & 1.11 & 2.18 & 2.72 & \textbf{1.11} & 2.35 & 0.40 & 1.11 & 2.76 & 2.83 & 2.57 & 4.09 \\
    MICE & 1.36 & 1.73 & 1.16 & 1.61 & 0.76 & \underline{0.93} & 2.09 & 2.25 & 2.04 & 2.27 & 0.59 & 1.22 & 2.43 & 2.57 & 3.32 & 4.00 \\
    SVD & 2.15 & 2.26 & 1.44 & 2.11 & 1.08 & 1.27 & 2.80 & 3.26 & 2.43 & 3.03 & 0.94 & 1.55 & 2.75 & 3.06 & 3.64 & 4.36 \\
    Spectral & 2.01 & 2.56 & 1.44 & 2.21 & 0.92 & 1.45 & 2.94 & 3.58 & 2.34 & 3.04 & 0.78 & 2.01 & 3.10 & 3.64 & 3.97 & 4.71 \\
    GAIN & 1.52 & 1.87 & 1.35 & 1.74 & 0.92 & 1.06 & 13.28 & 13.31 & 2.20 & 2.31 & 1.23 & 1.48 & 2.61 & 2.59 & 3.40 & 3.85 \\
    OT & 1.18 & 1.76 & 0.91 & 1.29 & 0.78 & 1.02 & 1.75 & 2.57 & 1.82 & \underline{2.15} & 0.44 & 0.85 & 2.57 & 2.58 & 2.50 & \textbf{3.60} \\
    Miracle & 1.30 & 1.90 & 1.11 & 2.50 & 0.73 & 1.66 & 1.88 & 4.89 & 2.03 & 4.23 & 0.62 & 2.86 & 2.43 & 3.31 & 3.62 & 7.50 \\
    GINN & 1.68 & \underline{1.56} & 1.63 & 1.69 & 1.17 & 0.96 & 1.80 & \underline{2.10} & 1.95 & \textbf{1.81} & 1.50 & 1.37 & 2.67 & 2.93 & 4.28 & 4.10 \\
    GRAPE & \underline{0.86} & 1.67 & \underline{0.76} & \underline{1.21} & \underline{0.63} & \underline{0.93} & \underline{1.61} & 2.35 & 1.55 & 2.23 & \underline{0.20} & \underline{0.50} & \underline{2.35} & \underline{2.51} & \underline{2.49} & 3.91 \\
    IGRM & \textbf{0.74} & \textbf{1.42} & \textbf{0.69} & \textbf{1.02} & \textbf{0.62} & \textbf{0.92} & \textbf{1.11} & \textbf{2.05} & \underline{1.51} & 2.24 & \textbf{0.18} & \textbf{0.27} & \textbf{2.27} & \textbf{2.41} & \textbf{2.41} & \underline{3.69} \\
    \bottomrule
    \end{tabular}%
    }
  \label{tab:imputationwithbaseline}%
\end{table*}%
\begin{table*}[htbp]
    \centering
    \caption{Ablation study. The time reported here is the initialization time(seconds) of friend network in Concrete.}
    \resizebox{\textwidth}{!}{
    \begin{tabular}{l|cccccccc|c}
    \toprule
    \textbf{Dataset} & \textbf{Concrete} & \textbf{Housing} & \textbf{Wine} & \textbf{Yacht} & \textbf{Heart} & \textbf{DOW30} & \textbf{E-commerce} & \textbf{Diabetes} & Time\\
    \midrule
    IGRM & \textbf{0.74}\footnotesize $\pm 0.01$ & \textbf{0.69}\footnotesize $\pm 0.01$ & \textbf{0.62}\footnotesize $\pm 0.00$ & 1.11\footnotesize $\pm 0.07$ & 1.51\footnotesize $\pm 0.01$ & \textbf{0.18}\footnotesize $\pm 0.01$ & 2.27\footnotesize $\pm 0.01$ & 2.41\footnotesize $\pm 0.02$ & 0.01 \\
    w/o GAE & 0.86\footnotesize $\pm 0.01$ & 0.80\footnotesize $\pm 0.02$ & 0.64\footnotesize $\pm 0.00$ & 1.63\footnotesize $\pm 0.06$ & 1.62\footnotesize $\pm 0.01$ & 0.20\footnotesize $\pm 0.00$ & 2.35\footnotesize $\pm 0.01$ & 2.68\footnotesize $\pm 0.04$ & 0.01\\
    once GAE & 0.80\footnotesize $\pm 0.01$ & 0.73\footnotesize $\pm 0.02$ & 0.63\footnotesize $\pm 0.01$ & 1.33\footnotesize $\pm 0.04$ & 1.57\footnotesize $\pm 0.07$ & 0.20\footnotesize $\pm 0.00$ & 2.31\footnotesize $\pm 0.02$ & 2.56\footnotesize $\pm 0.08$ & 0.01\\
    \midrule
    IGRM-cos & 0.76\footnotesize $\pm 0.01$ & 0.70\footnotesize $\pm 0.01$ & 0.63\footnotesize $\pm 0.01$ & 1.29\footnotesize $\pm 0.04$ & \textbf{1.50}\footnotesize $\pm 0.01$ & 0.19\footnotesize $\pm 0.01$ & 2.27\footnotesize $\pm 0.01$ & \textbf{2.38}\footnotesize $\pm 0.03$ & 86.50\\
    IGRM-rule & 0.76\footnotesize $\pm 0.00$ & \textbf{0.69}\footnotesize $\pm 0.01$ & 0.62\footnotesize $\pm 0.01$ & \textbf{1.09}\footnotesize $\pm 0.01$ & 1.55\footnotesize $\pm 0.03$ & 0.19\footnotesize $\pm 0.01$ & \textbf{2.27}\footnotesize $\pm 0.00$ & 2.40\footnotesize $\pm 0.04$ & 12.99\\
    \bottomrule
    \end{tabular}
    }
    \label{tab:ablation}
\end{table*}

\subsubsection{Parameter settings}
For all baselines, we follow their default settings from their papers. IGRM employs three variant GraphSAGE layers with 64 hidden units for bipartite GRL and one GraphSAGE layer for friend network GRL. The Adam optimizer with a learning rate of 0.001 and the ReLU activation function is used. 
In the process of initializing $\mathcal{F}$, we randomly connect samples with sample size $|\mathbf{U}|$ edges to build the initial friend network. This structure of the friend network is reconstructed per 100 epochs during the bipartite graph training. As the datasets are completely observed, we introduce certain percentages of missing data with missing with complete random mask(MCAR), missing at random (MAR), and missing not at random (MNAR) with different missing ratios. We treat the observed values as train set and missing values as the test set. Values are scaled into $[0,1]$ with the MinMax scaler~\cite{leskovec2020mining}.
For all experiments, we train IGRM for 20,000 epochs, run five trials with different seeds and report the mean of the mean absolute error(MAE). Due to page limits, results for MAR and MNAR with settings \cite{muzellec2020missing} are shown in Appendix. 


\subsection{Experimental Results}

Table~\ref{tab:imputationwithbaseline} shows the mean of MAE compared with various methods with 30\% and 70\% missing ratios, the best result is marked in bold, and the second is underlined. For 30\% missing, IGRM outperforms all baselines in 7 out of 8 datasets and reaches the second best performance in Heart, and its MAE is 2\%\textasciitilde 31\% lower compared with the best baseline GRAPE.
Mean imputation only uses a single dimension of features for imputation and reaches almost the worst results. 
SVD and Spectral highly depend on certain hyper-parameters, which results in erratic performance across different datasets, even worse results in 5 datasets than Mean. 
GAIN is also not stable. It has the most significant variance and gets the worst result in Yacht. 
GRAPE explicitly models the pure bipartite graph and learns highly expressed representations of observed data. Its performance is comparably stable and usually ranks second. However, it fails to get weights from similar samples. IGRM introduces a friend network based on the pure bipartite graph, encouraging information propagation between similar samples, which is beneficial for imputation. Furthermore, the performances of most baselines are worse than Mean under the 70\% missing ratio, while IGRM still maintains good performance.

\begin{figure}[bt]
    \centering
    \subfigbottomskip=2pt
\includegraphics[width=0.9\columnwidth]{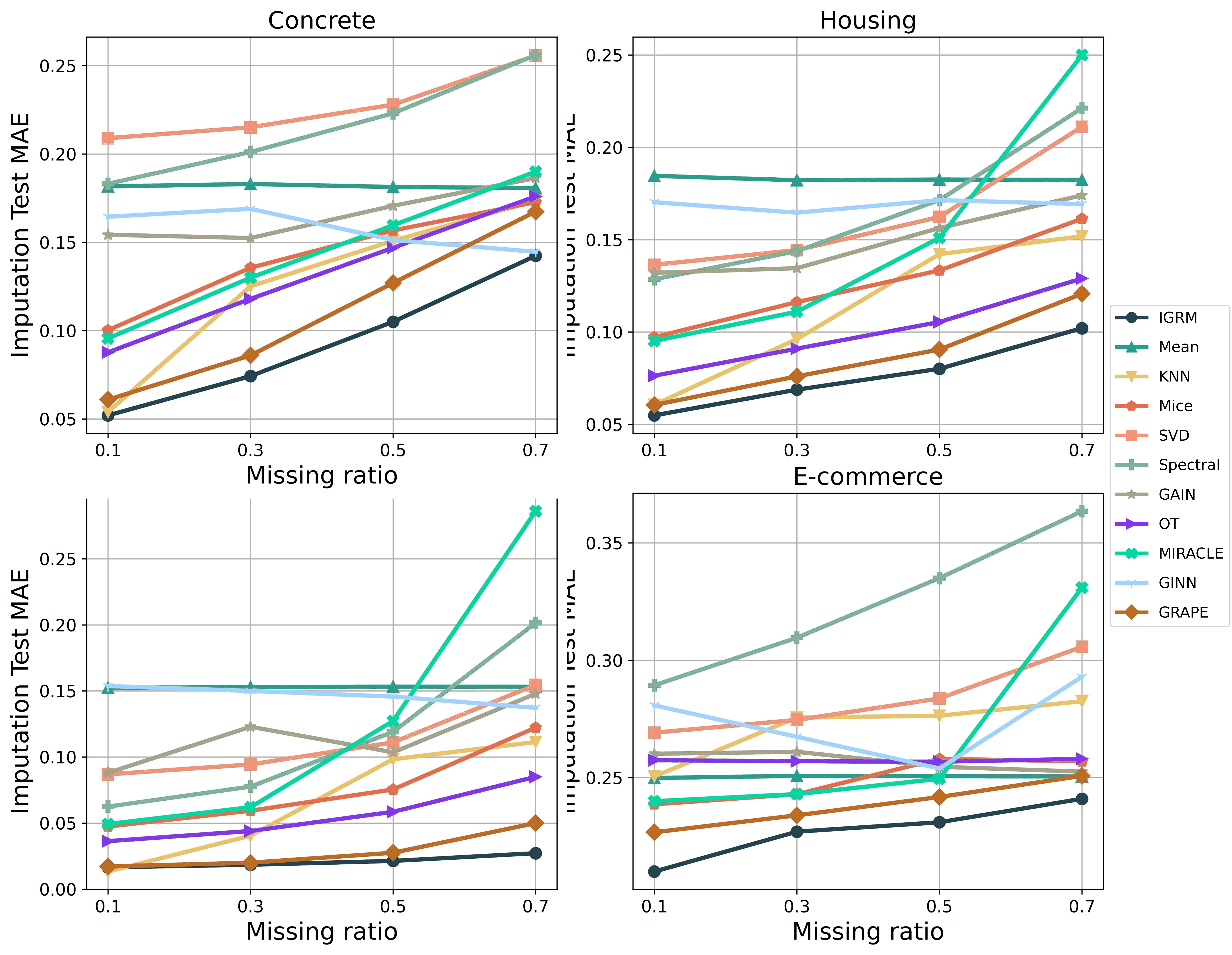}
    \caption{Averaged MAE of feature imputation with different missing ratios over five trials.}
    \label{fig:missingratio}
\end{figure}

\subsubsection{Different missing ratio}
To examine the robustness of IGRM with respect to the missing ratios of the data matrix: 0.1,0.3,0.5,0.7. The curves in Fig.\ref{fig:missingratio} demonstrate 
the trends of all methods. With the increase of the missing ratio, the performance of almost all algorithms except for Mean, generally decreases. In comparisons,  IGRM consistently outperforms other baselines across almost the entire range of missing ratios. In the missing ratio of 0.1, KNN also performs well. A possible explanation for these results is that low missing ratios do not significantly distort the clustering results. However, KNN suffers significant performance deterioration with a high missing ratio due to the large similarity deviation from missing data. While IGRM relieves the problem of data sparse by constructing the friend network.


\subsection{Embedding Evaluation}
In this section, we evaluate the qualities of generated sample embeddings from IGRM.

\subsubsection{Quality of embeddings}
\begin{figure}[tbp]
	\centering
	\subfigbottomskip=2pt
	\subfigure[IGRM-Yacht(0.17)]{
		\includegraphics[width=0.45\columnwidth]{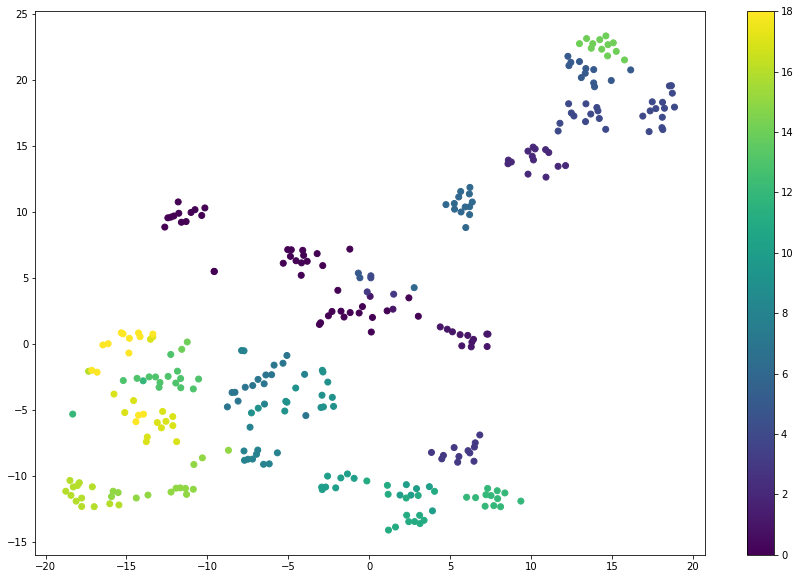}}
	\subfigure[GRAPE-Yacht(0.14)]{
		\includegraphics[width=0.46\columnwidth]{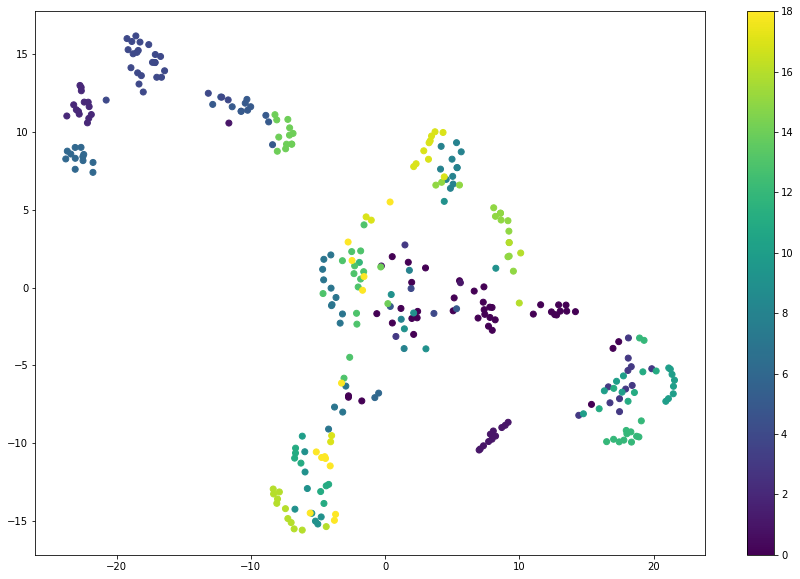}}
	\caption{The t-SNE embeddings of nodes from IGRM and GRAPE. The value in () indicates the silhouette coefficient.}
	\label{fig:tsne}
\end{figure}

We visualize sample embeddings of IGRM and GRAPE with t-SNE~\cite{2008Visualizing} and calculate the average silhouette score~\cite{rousseeuw1987silhouettes} for all sample embeddings. We compare them with the clusters generated with the ground-truth complete tabular data. The higher value of the silhouette score means embeddings are better separated. Fig.\ref{fig:tsne} shows the embeddings distribution in Yacht. We can observe that IGRM tends to be better clustered and have clearer boundaries between classes than GRAPE. IGRM has a silhouette score of 0.17, about 35.7\% higher than 0.14 in GRAPE. As a result, IGRM achieves a MAE 31\% lower than GRAPE.


\begin{figure}[tbp]
	\centering 
	\subfigbottomskip=2pt
	\subfigure[DOW30]{
		\includegraphics[width=0.45\columnwidth]{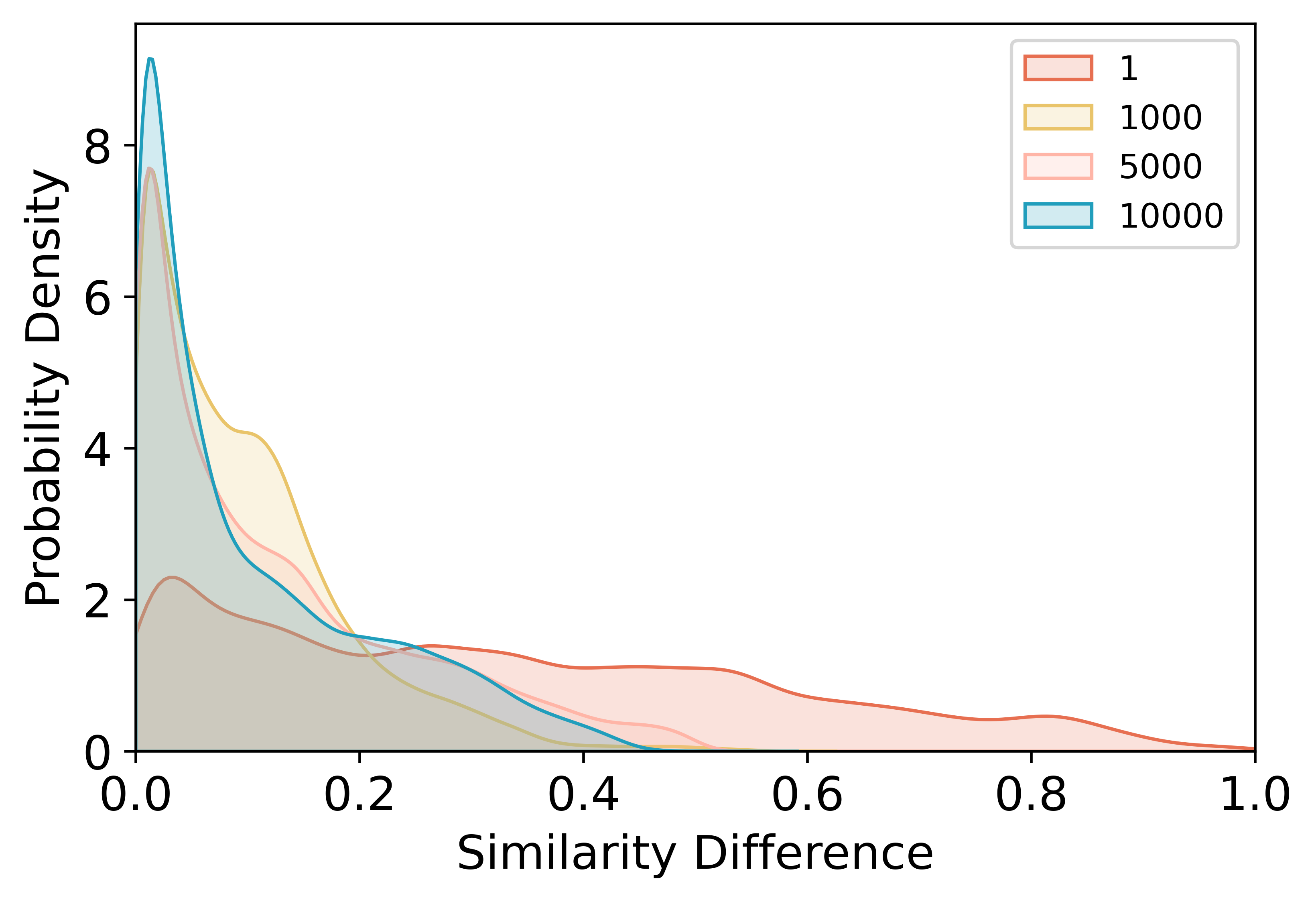}}
	\subfigure[Yacht]{
		\includegraphics[width=0.45\columnwidth]{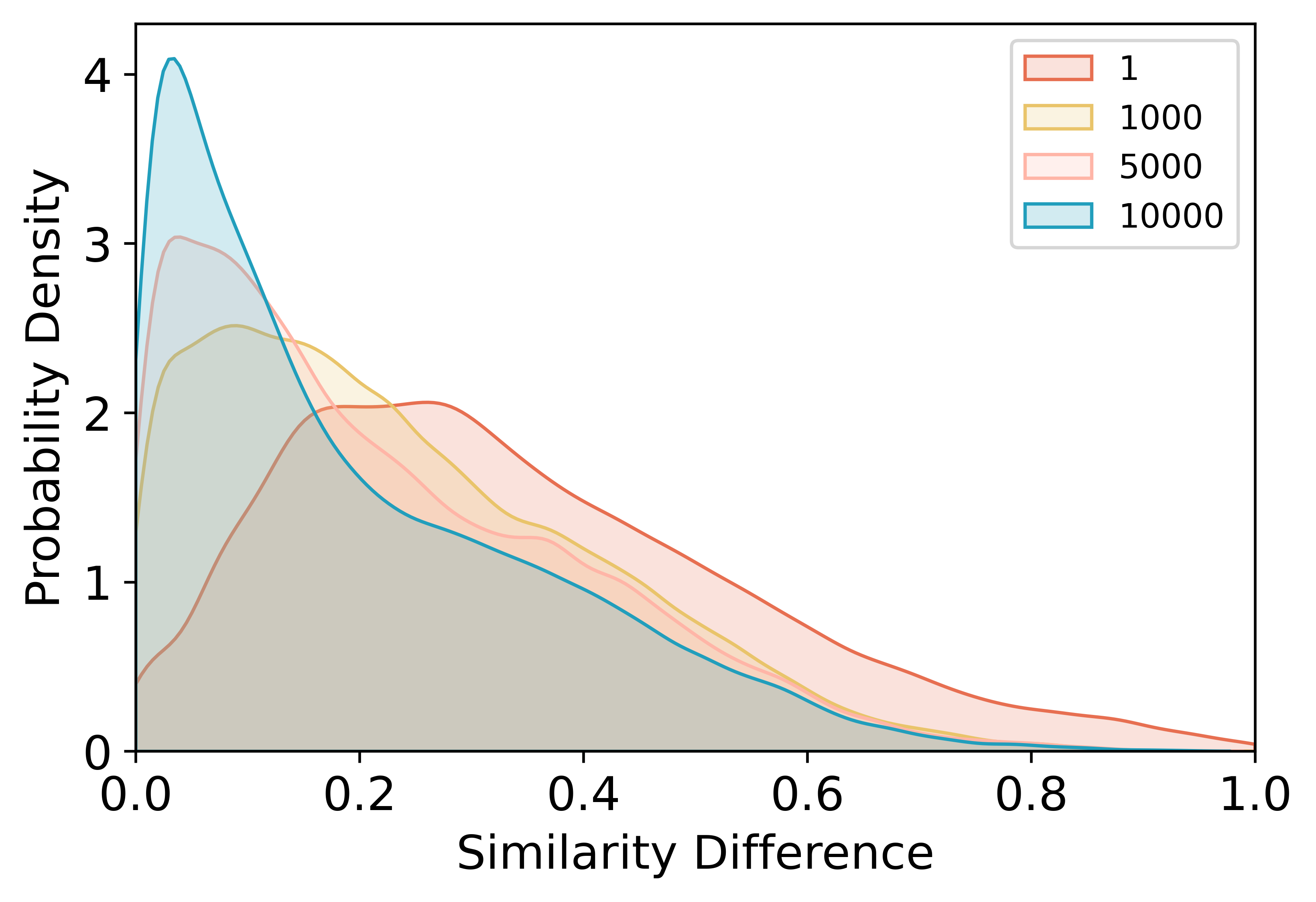}}
    \caption{Trends of similarity deviation distribution during training with missing ratio 0.3. The deviation is the absolute value of the difference between the generated embeddings similarity and the ground-truth similarity.}
	\label{fig:similarity_iterative}
\end{figure}

\subsubsection{Evolution of embeddings}
To prove the effectiveness of iterative friend network reconstruction, we show the cosine similarity deviation distribution at 1st, 1,000th, 5,000th, and 10,000th epoch in DOW30 and Yacht with missing ratio 0.3. Fig.\ref{fig:similarity_iterative} shows that the quality of generated embeddings gradually improves during the training process. The similarity deviation in the 1st epoch is relatively high with most distributed around 0.3$\sim$0.6 and a few less than 0.2. During training, the quality of the embeddings is much improved. The similarity of node embeddings at the 5,000th epoch is already highly close to the ground truth. It verifies the effectiveness of the iterative friend network reconstruction.

\subsection{Ablation Study}
In this subsection, we study the impact of different designs and show their impacts on the data imputation.

\subsubsection{Impacts from friend network reconstruction.} We test the performance with the friend network reconstruction component turned off(w/o GAE) or only reconstructing the friend network once at the first epoch(once GAE). The results are shown in Table.~\ref{tab:ablation}. IGRM(w/o GAE) suffers significant performance deterioration consistently on all datasets, sometimes even worse than GRAPE. We conjecture this is because the initial random-generated friend network is very noisy. Training on a random friend network without reconstruction can hardly helpful. With even one reconstruction, IGRM(once GAE) achieves significant performance improvements over IGRM(w/o GAE) across almost all tested datasets. This result clearly shows the effectiveness of friend network reconstruction. For the iterative reconstruction, almost all compared datasets show further improvements when the friend network is optimized iteratively, compared to IGRM(once GAE). 
Results demonstrate that the friend network at the early stage is inaccurate, and the iterative reconstruction is essential for IGRM.

\subsubsection{Impacts of friend network initialization.}
We further investigate how the friend network's initialization method(random, rule, cos) affects IGRM's performance. IGRM-cos calculates the exploits cosine similarity among samples and connects the most similar sample pairs with the same number of IGRM-random. IGRM-rule uses FP-growth~\cite{han2000mining} to mind frequent data patterns in samples and connects samples failing in the same rule. In IGRM-rule, we discrete the continuous data with Davies-Bouldin index~\cite{davies1979cluster} to handle continuous features. Detailed descriptions of the IGRM-rule can be found in Appendix. The lower part of Table.~\ref{tab:ablation} shows three methods have no significant performance differences while the random initialization is about four orders faster. The complexity of IGRM-rule is $\mathcal{O}(nlogn)$ and even higher $\mathcal{O}(n^2)$ for IGRM-cos. As the iterative reconstruction process can largely mitigate the importance of friend network initialization, IGRM uses random initialization as default. 

\subsubsection{Network reconstruction frequency}
One of the major overheads introduced by IGRM is the friend network reconstruction. Table.~\ref{tab:update} shows the effect of friend network reconstruction with different update frequencies(per 1/10/100 epochs). The speedup ratio is the time spent with reconstruction per 1 epoch divided by reconstruction with other frequencies. Results before / are time ratio between per 1 and 10 epochs and after are time ratio between per 1 and 100 epochs. With different update frequency, the imputation results largely unchanged. However, a large speed up is observed when per 100 epochs is used, especially for big datasets, e.g. E-commerce. Therefore, IGRM reconstructs the friend network per 100 epochs.
\begin{table}[htbp]
    \centering
    \caption{Results and speedup v.s network update frequency.}
    \resizebox{\columnwidth}{!}{
    \begin{tabular}{l|ccc|c}
    \toprule
    \textbf{Per \# Epoch} & \textbf{1} & \textbf{10} & \textbf{100} & \textbf{Speedup} \\
    \midrule
    Concrete & 0.76\footnotesize $\pm 0.01$ & \textbf{0.74}\footnotesize $\pm 0.01$ & 0.74\footnotesize $\pm 0.02$ & 4.36/6.53 \\
    Housing & 0.69\footnotesize $\pm 0.01$ & \textbf{0.68}\footnotesize $\pm 0.01$ & 0.69\footnotesize $\pm 0.01$ & 1.51/2.03 \\
    Wine & 0.62\footnotesize $\pm 0.01$ & \textbf{0.61}\footnotesize $\pm 0.00$ & 0.62\footnotesize $\pm 0.01$ & 4.16/6.24 \\
    Yacht & 1.12\footnotesize $\pm 0.09$ & \textbf{1.06}\footnotesize $\pm 0.07$ & 1.11\footnotesize $\pm 0.08$ & 1.71/1.92 \\
    Heart & 1.60\footnotesize $\pm 0.03$ & 1.52\footnotesize $\pm 0.03$ & \textbf{1.51}\footnotesize $\pm 0.03$ & 3.18/4.37 \\
    DOW30 & 0.19\footnotesize $\pm 0.00$ & 0.19\footnotesize $\pm 0.00$ & \textbf{0.18}\footnotesize $\pm 0.01$ & 5.37/9.40 \\
    E-commerce & 2.28\footnotesize $\pm 0.01$ & 2.27\footnotesize $\pm 0.01$ & \textbf{2.27}\footnotesize $\pm 0.00$ & 8.22/27.48 \\
    Diabetes & 2.42\footnotesize $\pm 0.03$ & 2.45\footnotesize $\pm 0.03$ & \textbf{2.41}\footnotesize $\pm 0.04$ & 1.93/2.14 \\
    \bottomrule
    \end{tabular}
     }
    \label{tab:update}
\end{table}





\section{Conclusion}
In this paper, we propose IGRM, an iterative graph generation and reconstruction based missing data imputation framework. First, we transform the missing data imputation problem into an edge-level prediction problem.
And under the assumption that similar nodes would have similar embeddings, we cast the similarity relations modeling problem in sample nodes as node embeddings similarity problem. IGRM jointly and iteratively learning both friend network structure and node embeddings optimized by and for the imputation task. Experimental results demonstrate the effectiveness and efficiency of the proposed model. In the future, we plan to explore effective techniques for handling the missing data problem in large datasets.

\section{Acknowledgement}
This work was supported by the National Science Foundation of China (61772473, 62073345 and 62011530148). The computing resources supporting this work were partially provided by High-Flyer AI. (Hangzhou High-Flyer AI Fundamental Research Co., Ltd.)

\bibliography{ref}

\begin{thebibliography}{42}
\providecommand{\natexlab}[1]{#1}

\bibitem[{Acuna and Rodriguez(2004)}]{acuna2004treatment}
Acuna, E.; and Rodriguez, C. 2004.
\newblock The treatment of missing values and its effect on classifier
  accuracy.
\newblock In \emph{Classification, clustering, and data mining applications},
  639--647. Springer.

\bibitem[{Allen and Li(2016)}]{allen2016generative}
Allen, A.; and Li, W. 2016.
\newblock Generative adversarial denoising autoencoder for face completion.
\newblock \emph{School of Interactive Computing, College of Computing, Georgia
  Institute of Technology}.

\bibitem[{Asuncion and Newman(2007)}]{asuncion2007uci}
Asuncion, A.; and Newman, D. 2007.
\newblock UCI machine learning repository.

\bibitem[{Azur et~al.(2011)Azur, Stuart, Frangakis, and
  Leaf}]{azur2011multiple}
Azur, M.~J.; Stuart, E.~A.; Frangakis, C.; and Leaf, P.~J. 2011.
\newblock Multiple imputation by chained equations: what is it and how does it
  work?
\newblock \emph{International journal of methods in psychiatric research},
  20(1): 40--49.

\bibitem[{Berg, Kipf, and Welling(2017)}]{berg2017graph}
Berg, R. v.~d.; Kipf, T.~N.; and Welling, M. 2017.
\newblock Graph convolutional matrix completion.
\newblock \emph{arXiv preprint arXiv:1706.02263}.

\bibitem[{Bertsimas, Pawlowski, and Zhuo(2017)}]{bertsimas2017predictive}
Bertsimas, D.; Pawlowski, C.; and Zhuo, Y.~D. 2017.
\newblock From predictive methods to missing data imputation: an optimization
  approach.
\newblock \emph{The Journal of Machine Learning Research}, 18(1): 7133--7171.

\bibitem[{Buuren and Groothuis-Oudshoorn(2010)}]{buuren2010mice}
Buuren, S.~v.; and Groothuis-Oudshoorn, K. 2010.
\newblock mice: Multivariate imputation by chained equations in R.
\newblock \emph{Journal of statistical software}, 1--68.

\bibitem[{Davies and Bouldin(1979)}]{davies1979cluster}
Davies, D.~L.; and Bouldin, D.~W. 1979.
\newblock A cluster separation measure.
\newblock \emph{IEEE transactions on pattern analysis and machine
  intelligence}, (2): 224--227.

\bibitem[{Donders et~al.(2006)Donders, Van Der~Heijden, Stijnen, and
  Moons}]{donders2006gentle}
Donders, A. R.~T.; Van Der~Heijden, G.~J.; Stijnen, T.; and Moons, K.~G. 2006.
\newblock A gentle introduction to imputation of missing values.
\newblock \emph{Journal of clinical epidemiology}, 59(10): 1087--1091.

\bibitem[{Fan, Zhang, and Udell(2020)}]{fan2020polynomial}
Fan, J.; Zhang, Y.; and Udell, M. 2020.
\newblock Polynomial matrix completion for missing data imputation and
  transductive learning.
\newblock In \emph{Proceedings of the AAAI Conference on Artificial
  Intelligence}, volume~34, 3842--3849.

\bibitem[{Genes et~al.(2016)Genes, Esnaola, Perlaza, Ochoa, and
  Coca}]{genes2016recovering}
Genes, C.; Esnaola, I.; Perlaza, S.~M.; Ochoa, L.~F.; and Coca, D. 2016.
\newblock Recovering missing data via matrix completion in electricity
  distribution systems.
\newblock In \emph{2016 IEEE 17th International Workshop on Signal Processing
  Advances in Wireless Communications (SPAWC)}, 1--6. IEEE.

\bibitem[{Gondara and Wang(2017)}]{gondara2017multiple}
Gondara, L.; and Wang, K. 2017.
\newblock Multiple imputation using deep denoising autoencoders.
\newblock \emph{arXiv preprint arXiv:1705.02737}, 280.

\bibitem[{Hamilton, Ying, and
  Leskovec(2017{\natexlab{a}})}]{hamilton_grapshsage}
Hamilton, W.; Ying, Z.; and Leskovec, J. 2017{\natexlab{a}}.
\newblock Inductive representation learning on large graphs.
\newblock In \emph{Advances in neural information processing systems},
  1024--1034.

\bibitem[{Hamilton, Ying, and
  Leskovec(2017{\natexlab{b}})}]{hamilton2017representation}
Hamilton, W.~L.; Ying, R.; and Leskovec, J. 2017{\natexlab{b}}.
\newblock Representation learning on graphs: Methods and applications.
\newblock \emph{arXiv preprint arXiv:1709.05584}.

\bibitem[{Han, Pei, and Yin(2000)}]{han2000mining}
Han, J.; Pei, J.; and Yin, Y. 2000.
\newblock Mining frequent patterns without candidate generation.
\newblock \emph{ACM sigmod record}, 29(2): 1--12.

\bibitem[{Hastie et~al.(2015)Hastie, Mazumder, Lee, and
  Zadeh}]{hastie2015matrix}
Hastie, T.; Mazumder, R.; Lee, J.~D.; and Zadeh, R. 2015.
\newblock Matrix completion and low-rank SVD via fast alternating least
  squares.
\newblock \emph{The Journal of Machine Learning Research}, 16(1): 3367--3402.

\bibitem[{J{\"a}ger, Allhorn, and Bie{\ss}mann(2021)}]{jager2021benchmark}
J{\"a}ger, S.; Allhorn, A.; and Bie{\ss}mann, F. 2021.
\newblock A benchmark for data imputation methods.
\newblock \emph{Frontiers in big Data}, 48.

\bibitem[{Jang, Gu, and Poole(2016)}]{jang2016categorical}
Jang, E.; Gu, S.; and Poole, B. 2016.
\newblock Categorical reparameterization with gumbel-softmax.
\newblock \emph{arXiv preprint arXiv:1611.01144}.

\bibitem[{Keerin, Kurutach, and Boongoen(2012)}]{keerin2012cluster}
Keerin, P.; Kurutach, W.; and Boongoen, T. 2012.
\newblock Cluster-based KNN missing value imputation for DNA microarray data.
\newblock In \emph{2012 IEEE International Conference on Systems, Man, and
  Cybernetics (SMC)}, 445--450. IEEE.

\bibitem[{Kipf and Welling(2016)}]{kipf2016variational}
Kipf, T.; and Welling, M. 2016.
\newblock Variational graph auto-encoders, 2016.
\newblock In \emph{Bayesian Deep Learning Workshop (NIPS 2016), arXiv preprint
  (arXiv: 161107308).[Google Scholar]}.

\bibitem[{Kusner and Hern{\'a}ndez-Lobato(2016)}]{kusner2016gans}
Kusner, M.~J.; and Hern{\'a}ndez-Lobato, J.~M. 2016.
\newblock Gans for sequences of discrete elements with the gumbel-softmax
  distribution.
\newblock \emph{arXiv preprint arXiv:1611.04051}.

\bibitem[{Kyono et~al.(2021)Kyono, Zhang, Bellot, and van~der
  Schaar}]{kyono2021miracle}
Kyono, T.; Zhang, Y.; Bellot, A.; and van~der Schaar, M. 2021.
\newblock MIRACLE: Causally-Aware Imputation via Learning Missing Data
  Mechanisms.
\newblock \emph{Advances in Neural Information Processing Systems}, 34:
  23806--23817.

\bibitem[{Laurens and Hinton(2008)}]{2008Visualizing}
Laurens, V. D.~M.; and Hinton, G. 2008.
\newblock Visualizing Data using t-SNE.
\newblock \emph{Journal of Machine Learning Research}, 9(2605): 2579--2605.

\bibitem[{Leskovec, Rajaraman, and Ullman(2020)}]{leskovec2020mining}
Leskovec, J.; Rajaraman, A.; and Ullman, J.~D. 2020.
\newblock \emph{Mining of massive data sets}.
\newblock Cambridge university press.

\bibitem[{Lin and Tsai(2020)}]{lin2020missing}
Lin, W.-C.; and Tsai, C.-F. 2020.
\newblock Missing value imputation: a review and analysis of the literature
  (2006--2017).
\newblock \emph{Artificial Intelligence Review}, 53(2): 1487--1509.

\bibitem[{Little and Rubin(1987)}]{little1987statistical}
Little, R.; and Rubin, D. 1987.
\newblock Statistical analysis with missing data.
\newblock Technical report, J. Wiley.

\bibitem[{Luo et~al.(2018)Luo, Cai, Zhang, Xu et~al.}]{luo2018multivariate}
Luo, Y.; Cai, X.; Zhang, Y.; Xu, J.; et~al. 2018.
\newblock Multivariate time series imputation with generative adversarial
  networks.
\newblock \emph{Advances in neural information processing systems}, 31.

\bibitem[{Maddison, Mnih, and Teh(2016)}]{maddison2016concrete}
Maddison, C.~J.; Mnih, A.; and Teh, Y.~W. 2016.
\newblock The concrete distribution: A continuous relaxation of discrete random
  variables.
\newblock \emph{arXiv preprint arXiv:1611.00712}.

\bibitem[{Malarvizhi and Thanamani(2012)}]{malarvizhi2012k}
Malarvizhi, M.~R.; and Thanamani, A.~S. 2012.
\newblock K-nearest neighbor in missing data imputation.
\newblock \emph{International Journal of Engineering Research and Development},
  5(1): 5--7.

\bibitem[{Mazumder, Hastie, and Tibshirani(2010)}]{mazumder2010spectral}
Mazumder, R.; Hastie, T.; and Tibshirani, R. 2010.
\newblock Spectral regularization algorithms for learning large incomplete
  matrices.
\newblock \emph{The Journal of Machine Learning Research}, 11: 2287--2322.

\bibitem[{Muzellec et~al.(2020)Muzellec, Josse, Boyer, and
  Cuturi}]{muzellec2020missing}
Muzellec, B.; Josse, J.; Boyer, C.; and Cuturi, M. 2020.
\newblock Missing data imputation using optimal transport.
\newblock In \emph{International Conference on Machine Learning}, 7130--7140.
  PMLR.

\bibitem[{Nazabal et~al.(2020)Nazabal, Olmos, Ghahramani, and
  Valera}]{nazabal2020handling}
Nazabal, A.; Olmos, P.~M.; Ghahramani, Z.; and Valera, I. 2020.
\newblock Handling incomplete heterogeneous data using vaes.
\newblock \emph{Pattern Recognition}, 107: 107501.

\bibitem[{Rousseeuw(1987)}]{rousseeuw1987silhouettes}
Rousseeuw, P.~J. 1987.
\newblock Silhouettes: a graphical aid to the interpretation and validation of
  cluster analysis.
\newblock \emph{Journal of computational and applied mathematics}, 20: 53--65.

\bibitem[{Spinelli, Scardapane, and Uncini(2020)}]{spinelli2020missing}
Spinelli, I.; Scardapane, S.; and Uncini, A. 2020.
\newblock Missing data imputation with adversarially-trained graph
  convolutional networks.
\newblock \emph{Neural Networks}, 129: 249--260.

\bibitem[{Troyanskaya et~al.(2001)Troyanskaya, Cantor, Sherlock, Brown, Hastie,
  Tibshirani, Botstein, and Altman}]{troyanskaya2001missing}
Troyanskaya, O.; Cantor, M.; Sherlock, G.; Brown, P.; Hastie, T.; Tibshirani,
  R.; Botstein, D.; and Altman, R.~B. 2001.
\newblock Missing value estimation methods for DNA microarrays.
\newblock \emph{Bioinformatics}, 17(6): 520--525.

\bibitem[{Vincent et~al.(2008)Vincent, Larochelle, Bengio, and
  Manzagol}]{vincent2008extracting}
Vincent, P.; Larochelle, H.; Bengio, Y.; and Manzagol, P.-A. 2008.
\newblock Extracting and composing robust features with denoising autoencoders.
\newblock In \emph{Proceedings of the 25th international conference on Machine
  learning}, 1096--1103.

\bibitem[{White, Royston, and Wood(2011)}]{white2011multiple}
White, I.~R.; Royston, P.; and Wood, A.~M. 2011.
\newblock Multiple imputation using chained equations: issues and guidance for
  practice.
\newblock \emph{Statistics in medicine}, 30(4): 377--399.

\bibitem[{Yoon, Jordon, and Schaar(2018)}]{yoon2018gain}
Yoon, J.; Jordon, J.; and Schaar, M. 2018.
\newblock Gain: Missing data imputation using generative adversarial nets.
\newblock In \emph{International Conference on Machine Learning}, 5689--5698.
  PMLR.

\bibitem[{You et~al.(2020)You, Ma, Ding, Kochenderfer, and
  Leskovec}]{you2020handling}
You, J.; Ma, X.; Ding, D.~Y.; Kochenderfer, M.; and Leskovec, J. 2020.
\newblock Handling missing data with graph representation learning.
\newblock \emph{arXiv preprint arXiv:2010.16418}.

\bibitem[{You et~al.(2019)You, Wu, Barrett, Ramanujan, and
  Leskovec}]{you2019g2sat}
You, J.; Wu, H.; Barrett, C.; Ramanujan, R.; and Leskovec, J. 2019.
\newblock G2SAT: Learning to generate sat formulas.
\newblock \emph{Advances in neural information processing systems}, 32.

\bibitem[{Zhang and Chen(2019)}]{zhang2019inductive}
Zhang, M.; and Chen, Y. 2019.
\newblock Inductive matrix completion based on graph neural networks.
\newblock \emph{arXiv preprint arXiv:1904.12058}.

\bibitem[{Zhao et~al.(2021)Zhao, Liu, Neves, Woodford, Jiang, and
  Shah}]{zhao2021data}
Zhao, T.; Liu, Y.; Neves, L.; Woodford, O.; Jiang, M.; and Shah, N. 2021.
\newblock Data augmentation for graph neural networks.
\newblock In \emph{Proceedings of the AAAI Conference on Artificial
  Intelligence}, volume~35, 11015--11023.

\end{thebibliography}


\end{document}